\documentclass{article}

\usepackage{PRIMEarxiv}

\usepackage[utf8]{inputenc} 
\usepackage[T1]{fontenc}    
\usepackage{amsmath}        
\usepackage{hyperref}       
\usepackage{url}            
\usepackage{booktabs}       
\usepackage{amsfonts}       
\usepackage{nicefrac}       
\usepackage{microtype}      
\usepackage{lipsum}
\usepackage{natbib}
\usepackage{fancyhdr}       
\usepackage{graphicx}       
\graphicspath{{media/}}     

\usepackage{graphicx}
\usepackage{subcaption}
\usepackage{enumerate}

\title{Beyond Reproducibility: Token Probabilities Expose Large Language Model Nondeterminism}

\author{
  Tairan Fu \\
  Politecnico di Milano \\
  Milano, Italy \\
   \small  \texttt{tairan.fu@polimi.it} \\
  \And
  Gonzalo Martínez, Javier Conde, Carlos Arriaga, Pedro Reviriego \\
  Information Processing and Telecommunications Center \\
  ETSI de Telecomunicación, Universidad Politécnica de Madrid \\
  Madrid, Spain \\
  \small \texttt{\{gonzalo.martinez.ruizdearcaute,javier.conde.diaz,carlos.arriaga.prieto,
  pedro.reviriego\}@upm.es} \\
\And
  Xiuyuan Qi, Shanshan Liu\\
  University of Electronic Science and Technology of China \\
  Chengdu, China \\
  \small \texttt{\{xiuyuanqi,ssliu\}@std.uestc.edu.cn} \\
}

\begin{document}
\maketitle

\begin{abstract}
The execution of Large Language Models (LLMs) has been shown to produce nondeterministic results when run on Graphics Processing Units (GPUs), even when they are configured to produce deterministic results, for example, by setting the temperature to zero. This is due to the finite precision effects of the arithmetic operations, which depend on the order in which they are executed. This order, in turn, depends on the processes that are running concurrently on the GPU. Previous studies have focused on the impact of nondeterminism on the text generated by the LLMs or on proposing mechanisms to achieve deterministic execution. This work takes a closer look at nondeterminism by analyzing the variations on the token probabilities, not on the generated text. This provides additional information that enables a deeper analysis. Interestingly, all the models evaluated have similar results in both the trends and the actual values of the variations of the probabilities. In particular, the results show that the effects of nondeterminism are significant for token probabilities that are in the range of 0.1 to 0.9, while they are much smaller when the probabilities are close to 0 or 1. Variations are largely independent of the magnitude of the probability, provided that they are in the 0.1 to 0.9 range. This has significant implications for our understanding of nondeterminism. The first is that nondeterminism will likely have a non-negligible impact on generated text when the temperature is not zero, as it introduces significant variations in the token probabilities except when they are close to 0 or 1. Secondly, it suggests that all models have similar non deterministic variations at the token probability level. Therefore, different variations in the performance of the generated text, for example, when measuring accuracy on a benchmark, seem to come from different token probabilities or response lengths. In the first case, models that have more token probabilities close to 0 or 1 will be less affected by nondeterminism, while in the case of length, models that produce shorter answers are also potentially less likely to be affected by nondeterminism. A third implication is that we may be able to estimate the impact of nondeterminism by running a single inference and analyzing the token level probabilities, instead of having to run the same inference many times. Finally, the data produced in the evaluation is shared in a public repository to facilitate further analysis by other researchers, which may lead to additional observations.
\end{abstract}


\section{Introduction}
\label{sec:Introduction}

The adoption of large language models (LLMs) has grown rapidly driven by their ability to perform a wide range of natural language processing tasks, including text generation, summarization, question answering, and code completion \cite{brown2020language, openai2023gpt4}. To perform inference, LLMs are commonly deployed on Graphic Processing Units (GPUs), which provide massive parallelism and optimized hardware for the tensor operations achieving the high performance needed to generate text at reasonable speed \cite{shoeybi2019megatron}. Major vendors supplying GPUs for LLM workloads 
include NVIDIA, which offers professional GPUs such as the A100 and H200, as well as high-end consumer GPUs like the RTX A6000, capable of supporting large-scale language model inference, and Huawei, with their Ascend series designed for AI workloads. These hardware platforms have become the standard backbone for deployment of LLMs, enabling models with billions to trillions of parameters to perform inference efficiently.

To maximize throughput, LLMs are often run with multiple concurrent inferences on the same GPU, a practice commonly referred to as batching. By processing several prompts simultaneously, the GPU can better use its massive parallelism more efficiently, utilizing tensor cores and memory bandwidth to perform operations across multiple sequences in parallel \cite{flexgen2023high},\cite{li2024large}. Batching improves utilization of GPU resources and reduces per‑prompt overhead, enabling faster response times in production environments and making it a widely adopted strategy for handling high query volumes with efficient resource usage. When multiple prompts are processed together in a batch, the GPU executes operations across sequences in parallel. The exact order in which computations are carried out can vary depending on thread scheduling, memory access patterns, or kernel optimizations.

The evaluation of large language models (LLMs) is essential for understanding their capabilities, limitations, and suitability for real‑world applications. As these models are increasingly used in domains such as healthcare, education, and finance, systematic assessment of metrics like accuracy, coherence, relevance, and safety has become a central concern in AI research \cite{zhang2025evaluating}. Without rigorous evaluation, apparent advances may mask subtle limitations in reasoning, factuality, or domain‑specific understanding, hampering the effective and safe integration of LLMs into practical systems \cite{laskar2024review}.

In the evaluation of large language models (LLMs), reproducible results are often required \cite{biderman2024lessons}. To ensure reproducible results, it is commonly assumed that setting the temperature to zero and fixing random seeds is enough. However, recent research shows that LLMs can remain non deterministic even in this case \cite{blackwell2024towards}. For example, \cite{atil2024non} provides a large-scale empirical study of this phenomenon, showing that inference runs with these settings can produce different outputs. The differences are attributed to floating-point rounding errors \cite{goldberg1991every} and non-deterministic kernel behaviors, especially in fused attention and normalization operations.  The role of floating-point non-associativity as a fundamental source of nondeterminism in high-performance computing and deep learning applications was studied in \cite{shanmugavelu2024impacts}, showing that mathematically identical computations can yield different results due to different orders in the operations. To improve reproducibility, the use of higher precision-floating point formats was proposed in \cite{yuan2025give}, showing that more stable results can be obtained when using them both for the parameters and arithmetic operations or only for the latter. However, this comes at the cost of, for example, doubling the memory because now each parameter needs 32 bits instead of 16 bits, and the arithmetic operations are also more complex. 

Nondeterminism is not only an issue for reproducibility; it can also be exploited to extract information from LLM providers as demonstrated in \cite{cai2025you} which relies on deterministic output comparisons to detect if a model has been changed to a lower cost or lower precision version. Potentially, an attacker could use nondeterminism to extract information about the hardware, batch size, or load of the provider. In fact, nondeterminism can be more broadly considered as a shift in the model's estimated token probabilities that has many potential implications beyond reproducibility, for example, it may affect performance when the temperature is larger than zero. 

Recent studies have primarily focused on observing differences in output sequences across inference runs, but less attention has been paid to token-level probability distributions and their variability. In this paper, a more general approach to the issue of nondeterminism is pursued, considering changes at the token-level probability rather than directly on the LLM generated text. Measuring the logits or log-probabilities of each token across repeated runs provides a more fine-grained view of nondeterminism, revealing effects that may not immediately result in a token change. Metrics such as the standard deviation and range of token probabilities variations across runs allow quantification of both typical fluctuations and extreme deviations, highlighting which tokens or probability regions are most sensitive to nondeterministic effects. By examining these measures across token sequences, it is possible to identify sensitive regions in the probability distribution and assess their potential impact on text outputs or downstream task performance. This approach complements previous work by providing a systematic framework for understanding per-step nondeterministic perturbations in LLMs, informing both reproducibility practices and robust evaluation. This enables a finer analysis of the effects of nondeterminism and leads to several contributions that improve our understanding of the nondeterminism of LLMs. In particular, this paper makes the following contributions:

\begin{enumerate}[a)]
    \item The nondeterminism is characterized as an uncertainty in the token probabilities, and two corresponding metrics to measure it are proposed.
    \item The effect of nondeterminism on Huawei GPUs is evaluated and reported for the first time.
    \item The uncertainty in the token probabilities is measured for different models, configurations, and hardware platforms, when the models run on a single GPU. 
    \item Based on the evaluation results, the following key trends are obtained: i) at the token probability level, the impact of nondeterminism is similar for all the models and GPUs considered; ii) the impact is significant when the token probability is in the range of 0.1 to 0.9 while much smaller when the probability is close to 0 or 1; iii) when the probability is not close to 0 or 1, the impact is similar regardless of the value of the probability (e.g., either 0.2 or 0.8).
    \item The implications of all results are discussed to illustrate the benefits of conducting the analysis at the token probability level.
    \item A public dataset containing the results is provided to facilitate further research.
\end{enumerate}

The rest of the paper is organized as follows: section \ref{sec:related work} summarizes the related work that has explored nondeterminism in LLMs; section \ref{sec:TokenProbs} introduces the metrics to measure variations in token probabilities due to nondeterminism; section \ref{sec:Methodology} presents the methodology used in the evaluation, the results are presented in \ref{sec:Evaluation} and discussed in section \ref{sec:Discussion} which also covers the limitations of our study. The paper ends with the conclusion in section \ref{sec:Conclusion}.

\section{Related Work}
\label{sec:related work}

Nondeterminism in large language models (LLMs) has been been recently studied, challenging the common assumption that setting the temperature to zero, fixing random seeds, and using single-batch inference guarantees consistent outputs. Empirical studies have shown that even under these intended “deterministic” configurations, repeated runs of the same prompt can produce different outputs \cite{atil2024non}. These variations are primarily attributed to small and subtle computational effects, including rounding differences in floating-point arithmetic and non-deterministic behaviors in GPU kernels, particularly in operations such as fused attention and normalization layers, which may execute operations in different orders depending on thread scheduling and low-level optimizations. The consequences of such nondeterminism are important. At the token level, it can induce small differences in logits and probabilities, which may affect top‑k token selections, sampled sequences, or subsequent decoding behavior when temperature is non-zero. Although individual variations are often small, their effects can propagate through the generation process, particularly in long sequences, resulting in outputs that differ in content, phrasing, or factual consistency. 

In addition to model settings, the impact of data formats has also been investigated. Commonly, several floating-point formats are used to run LLMs, including the traditional FP16 and FP32, defined in the IEEE-754 Standard \cite{ieee754-2019} as half and single precision formats with 16 and 32 bits, respectively, and the newer BF16 \cite{intel2018bf16} which is commonly used for models running in commercial GPUs. There are also efforts to develop emerging floating-point formats with fewer bits to reduce the memory and complexity of arithmetic operations \cite{micikevicius2022fp8} and circuits that combine multiplication and addition to reduce rounding errors \cite{Farzad}.  Previously, the role of floating-point non-associativity \cite{goldberg1991every} as a fundamental source of nondeterminism in high-performance computing and deep learning has been explored in \cite{shanmugavelu2024impacts}. The study highlights that mathematically identical computations can produce different results depending on the order of operations, which directly impacts the reproducibility of deep learning models. Therefore, to mitigate such a reproducibility problem in LLMs, \cite{yuan2025give} investigates the use of higher-precision floating-point formats, showing that FP32 arithmetic applied to parameters, computation, or both can reduce nondeterministic fluctuations compared to BF16 and FP16, albeit at the cost of increased memory and computational overhead. 

Beyond concerns about reproducibility, nondeterminism in large language models (LLMs) can have significant security and auditing implications. As demonstrated in \cite{cai2025you}, nondeterministic outputs can be leveraged to detect subtle changes in model deployments, such as when a provider substitutes a higher-cost or higher-precision model with a lower cost or lower precision variant. By comparing outputs over repeated queries, an attacker or auditor may be able to infer differences in hardware configurations, batch sizes, or server load conditions, potentially exposing internal operational details of the LLM service. These findings suggest that nondeterminism is not merely a technical nuisance but could be exploited to extract sensitive information or detect optimization strategies applied by model providers. Consequently, understanding and characterizing nondeterministic effects is important not only for ensuring scientific reproducibility, but also for security auditing, regulatory compliance, and trustworthiness in environments where LLM outputs are used in sensitive decision-making or deployed as APIs.

In addition to the works on nondeterminism in LLMs, several studies have examined stability, variability, and reproducibility in deep neural networks more broadly. For example, \cite{shanmugavelu2024impacts} shows that floating-point non-associativity and parallel GPU/kernel execution can cause significant variability across runs, even when code, hyperparameters, and random seeds are held constant. Similarly, \cite{pinto2022reproducibility} demonstrates that training the same convolutional neural network multiple times under identical settings can lead to substantial differences in learned parameters and outputs, especially when GPU operations are non-deterministic, and that using higher precision arithmetic reduces, but does not eliminate this variability. Finally, \cite{kim2023generative} investigates reproducibility in generative models in general beyond LLMs, showing that even under deterministic decoding strategies, reproducibility is not always guaranteed, and that consensus or multiple independent runs may be needed to verify results.

These results together show that nondeterminism in deep learning can stem from floating-point arithmetic effects, hardware/kernel implementations, and runtime execution conditions during inference. Nonetheless, most prior work has focused on model level reproducibility or output level consensus, rather than the impact of nondeterminism on the internal token probability distributions during typical inference. This gap motivates a more fine grained analysis of nondeterminism at the token level of probability, as pursued in this paper. This analysis at the token probability level, is a first step toward understanding nondeterminism impact on text generation under normal operating conditions, i.e., when the temperature is greater than zero.

Nondeterministic effects in large language models can vary across GPU vendors, architectures, and model implementations. Most existing studies have focused on NVIDIA GPUs. Differences in hardware, such as tensor cores, memory hierarchies, and execution pipelines, can influence the magnitude and characteristics of these variations. While NVIDIA hardware has been extensively benchmarked, alternative accelerators such as Huawei’s Ascend series are increasingly used for LLM inference. However, the investigation of nondeterminism on these platforms remain largely unexplored, leaving a gap in understanding how vendor-specific hardware and software stacks contribute to variations in LLM outputs. In this paper, an initial step is made in this direction by evaluating for the first time the impact of nondeterminism on Huawei GPUs.

\section{Token Probability-based Nondeterminism Metrics}
\label{sec:TokenProbs}

Most of the existing works have focused on the differences in the generated text due to nondeterminism when the LLMs are run with the temperature set to zero. However, this provides only indirect information on the variability of LLMs. Looking at the token probabilities corresponding to a temperature equal to one provides additional information. In fact, the changes in the generated text are a consequence of the changes in the token probabilities. 


Consider an inference step that generates a token selection from a vocabulary $v(j)$ with $j=1,2,...V$. Then, for each token $t$ in a given run $i$ we have the following probabilities on the vocabulary $p(v(j))_{i}$. So, over $N$ runs, we can consider each probability $p(v(j))$ as a random variable and compute the standard deviation of the $N$ samples, each corresponding to a run. Ideally, it should be zero, which corresponds to deterministic execution.

\[
\sigma_{\text{j}} = 
\sqrt{
\left[
\frac{1}{N} \sum_{i=1}^{N} 
\left( p(v(j))_i - \overline{p(v(j))} \right)^2
\right]
}
\]

The standard deviation of Eq. (1) provides useful information, but it is also interesting to look at the extremes of the distributions. To do so, we can compute the range of probabilities by computing the maximum and minimum on the $N$ runs and taking the difference as follows:

\[
R_{j} = \max_{1 \le i \le N}  p(v(j))_i -  \min_{1 \le i \le N}  p(v(j))_i.
\]

The standard deviation $\sigma_i$ and the range $R_i$ provide information on the impact of nondeterminism at each inference step on each possible token $v(j)$. Therefore, they are proposed as the evaluation metrics used in the rest of the paper to quantify the variations induced by nondeterminism.

\section{Evaluation Methodology}
\label{sec:Methodology}

\begin{figure}[htbp] 
    \centering
    \includegraphics[width=0.55\linewidth]{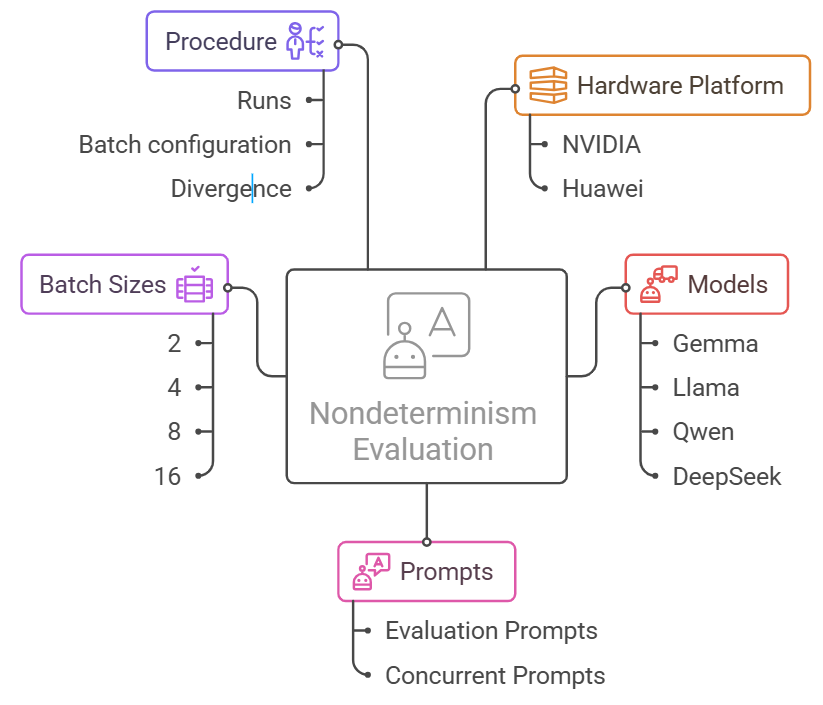}
    \caption{Diagram showing the factors involved in nondeterminism evaluation.}
    \label{fig:non-determinism-factors}
\end{figure}

To evaluate nondeterminism we must select the models to evaluate, the hardware platform to run them, the prompts and the batch sizes, and the procedure to use as illustrated in Figure \ref{fig:non-determinism-factors}. The subsections next discuss each of these evaluation parameters and procedures in detail.

\subsection{Models}

The evaluation focuses on models that can run on a single GPU. The rationale for this decision is twofold. Firstly, this should be the best case for deterministic behavior, as running the models on several devices is likely to introduce additional variations during execution. Therefore, it provides a lower bound for nondeterminism. Secondly, it corresponds to a common scenario for local deployment of LLMs, making the results widely applicable. 

From the models available at the time of writing this paper, we selected four LLMs with competitive performance from different companies so that the set is diverse and represents the state-of-the-art in models that can be run on a single GPU. Three are general models and the fourth is a reasoning model derived from another of the models evaluated. This enables us to evaluate whether reasoning has any significant impact on nondeterminism.

The selected models are the following. 

\begin{enumerate}[a)]
    \item Gemma3-12B: model from Google\footnote{\url{https://huggingface.co/google/gemma-3-12b-it}}. 
    \item Llama3.2-11B  model from Meta\footnote{\url{https://huggingface.co/meta-llama/Llama-3.2-11B-Vision-Instruct}}.
    \item Qwen3-VL-8B  model from Alibaba\footnote{\url{https://huggingface.co/Qwen/Qwen3-VL-8B-Instruct}}.
    \item DeepSeek-Qwen3-8B  model from DeepSeek\footnote{\url{https://huggingface.co/deepseek-ai/DeepSeek-R1-0528-Qwen3-8B}}.
\end{enumerate}

\subsection{Hardware platforms}

For our experiments, we used industry‑standard high‑performance GPUs. In particular, we considered configurations based on several NVIDIA GPUs: the A100 \cite{nvidia-a100-datasheet} data-center GPU, the consumer RTX-A6000 GPU \cite{nvidia-a6000-datasheet}, and the latest data‑center GPU H200 \cite{nvidia-h200-datasheet}. We also evaluate alternative AI accelerators such as Huawei Ascend-910 \cite{huawei-ascend910}, which also support LLM inference.

\subsection{Batch sizes}

The batch size determines the number of input prompts that a model simultaneously processes during a forward pass. Larger batch sizes can increase throughput, but may affect memory usage and, in some cases, computation order \cite{YU2023100151}. The batch size is the factor that will induce nondeterminism when a model runs on a single GPU; this is valid because the prompt under evaluation will run concurrently with other prompts which can lead to different order in the operations. To keep the computational effort reasonable, we use a logarithm scale to select batch sizes and consider the following values in the evaluation: 2, 4, 8, 16. These values should show the trends of nondeterminism with batch size if they exist.   

\subsection{Prompts}

In experiments, we have two types of prompts: 

\begin{enumerate}[a)]
    \item \textbf{Evaluation prompts}: the prompts on which we will assess nondeterminism.
    \item \textbf{Concurrent prompts}: the prompts that will run concurrently with the evaluation prompts when the batch size is larger than one.
\end{enumerate}

Since the goal is to evaluate nondeterminism in realistic scenarios, we decided to use a homogeneous set of prompts to load the system and also partly for the evaluation prompts. To do so, we take questions from the Massive Multitask Language Understanding (MMLU) dataset \cite{hendrycks2020measuring} which covers 57 diverse different categories ranging from algebra to world religions. This ensures that the questions are diverse and, at the same time, facilitates the reproducibility of our experiments and conducting additional experiments. In more detail, for each batch of size $B$, we randomly select $B-1$ questions from the more than 15,000 questions in MMLU and ask the model to answer the question providing an explanation. Therefore, the model answers questions on a wide range of topics that mimic the use of models in many real scenarios. 

For the evaluation prompts, we also use randomly selected questions from the MMLU dataset as the ones used for the concurrent prompts. The idea is to test prompts that are similar to the rest of the prompts being executed in the system and that cover a wide range of topics. 


\subsection{Testing procedure}

To evaluate nondeterminism, we set the batch size to $B$ and construct batches using $B-1$ randomly selected MMLU questions and one evaluation prompt. In more detail:

\begin{enumerate}[a)]
    \item Select 10 questions randomly from MMLU. These will be the set of 10 evaluation prompts.
    \item Select $B-1$ questions from MMLU.
    \item Run ten batches, one per evaluation prompt using the $B-1$ prompts selected in 2) and one of the evaluation prompts.
\end{enumerate}

The analysis is then conducted up to the point where at least one of the runs diverges, ensuring that all tokens included in the comparison were generated based on exactly the same preceding context. This approach guarantees a fair and consistent evaluation of token-level variations across runs.

To adequately characterize the distribution of nondeterministic effects in model outputs, we conduct $N$ independent runs. The standard error of the estimated standard deviation is approximately
\[
SE_\text{std} \approx \frac{\sigma}{\sqrt{2(N-1)}}.
\] 
This expression indicates that the precision the standard deviation scales with $1/\sqrt{N}$, independent of the magnitude of the underlying variability \cite{montgomery2014applied}. In addition, the observed range of token probabilities across runs provides a complementary measure of variability that captures the maximum spread of probabilities due to nondeterminism. In all the experiments, we set $N = 50$ to ensure that both the standard deviation, and range are estimated with sufficient precision for tokens across different probability values, offering a practical compromise between statistical reliability and computational cost.

\section{Evaluation results}
\label{sec:Evaluation}

In this section, the results of the experiments are presented and discussed. First, we look at the distributions of the ranges $R_i$ and standard deviation $\sigma_i$ (i.e., the two proposed metrics) of the differences due to nondeterminism and of the token probabilities. Then, we analyze the results for each token probability value to understand the effects of the different parameters, such as model, batch size, or hardware platform. Finally, an additional experiment is conducted to evaluate whether model size has any relevant effect on nondeterminism. The code and the results of all the experiments are available in a public repository \footnote{\url{https://github.com/aMa2210/llm-nondeterminism/}}.  

\subsection{Distributions of token probability variations: range and standard deviation}

\begin{figure}[ht]
    \centering
    \includegraphics[width=0.98\textwidth]{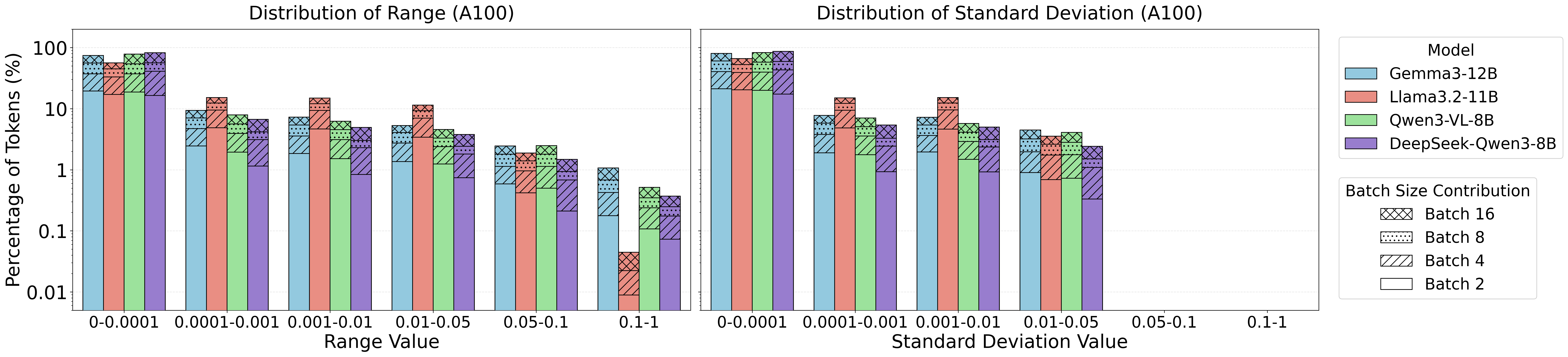}
    \includegraphics[width=0.98\textwidth]{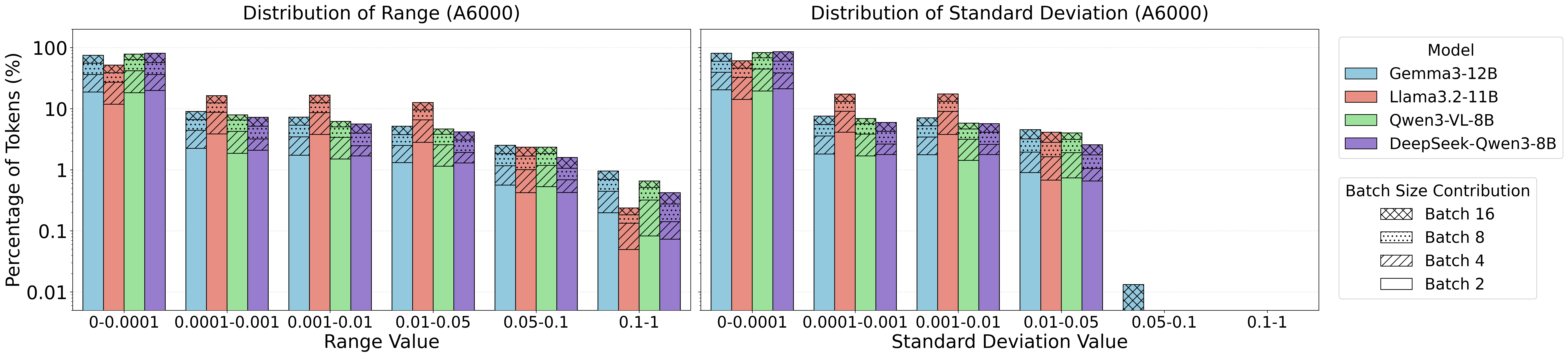}
    \includegraphics[width=0.98\textwidth]{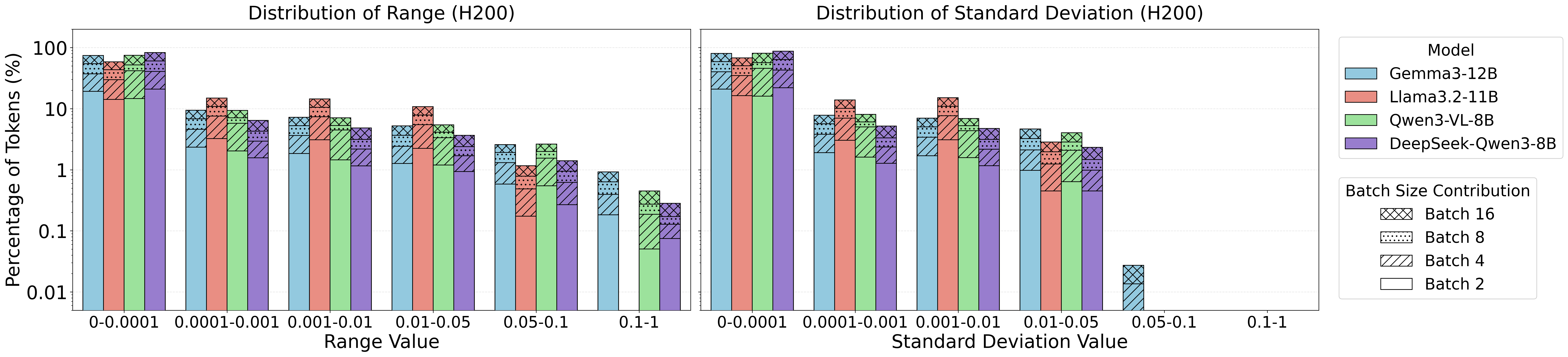}
    \includegraphics[width=0.98\textwidth]{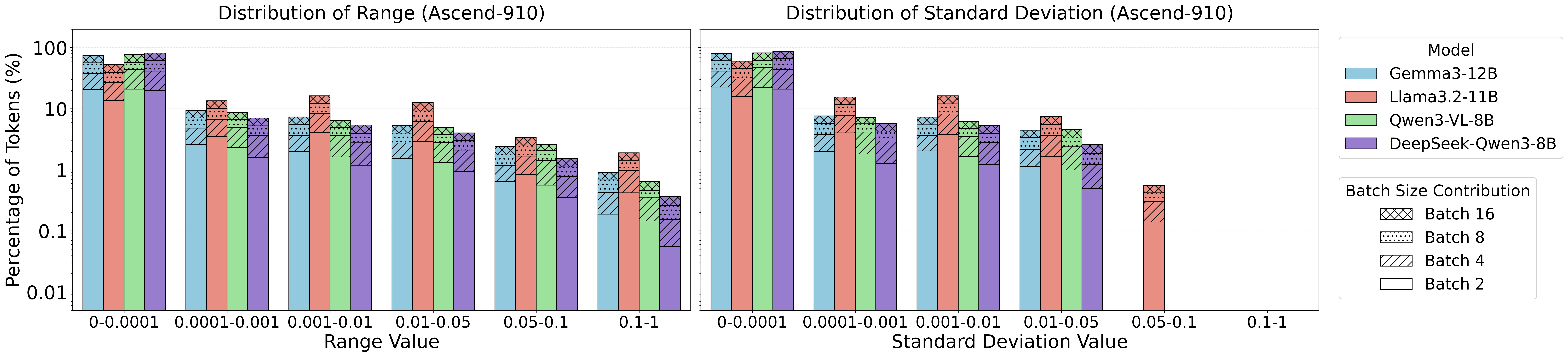}
    \caption{Distribution of average range $R_j$ and standard deviation $\sigma_j$ of variations in token probability due to nondeterminism for different batch sizes and models across different GPUs: NVIDIA A100 (top), NVIDIA A6000, NVIDIA H200, Huawei Ascend-910}
    \label{fig:range-SD-histogram}
\end{figure}

The results in terms of the distribution of the range and standard deviation of variations due to nondeterminism of the top-10 token probabilities across the 50 runs for all generated tokens are shown in Figure \ref{fig:range-SD-histogram}.
It can be seen that in all the models evaluated, for most tokens the values of the variations due to nondeterminism are close to zero with approximately 90\% having a range less than 0.01\% variations over the 50 runs and only roughly 1\% reaching 5\%. This seems to suggest that the effects of nondeterminism are low. It is also interesting to note that the trends and values are similar for all models both for range and standard deviation.  

\begin{figure}[ht]
    \centering
    \includegraphics[width=0.98\textwidth]{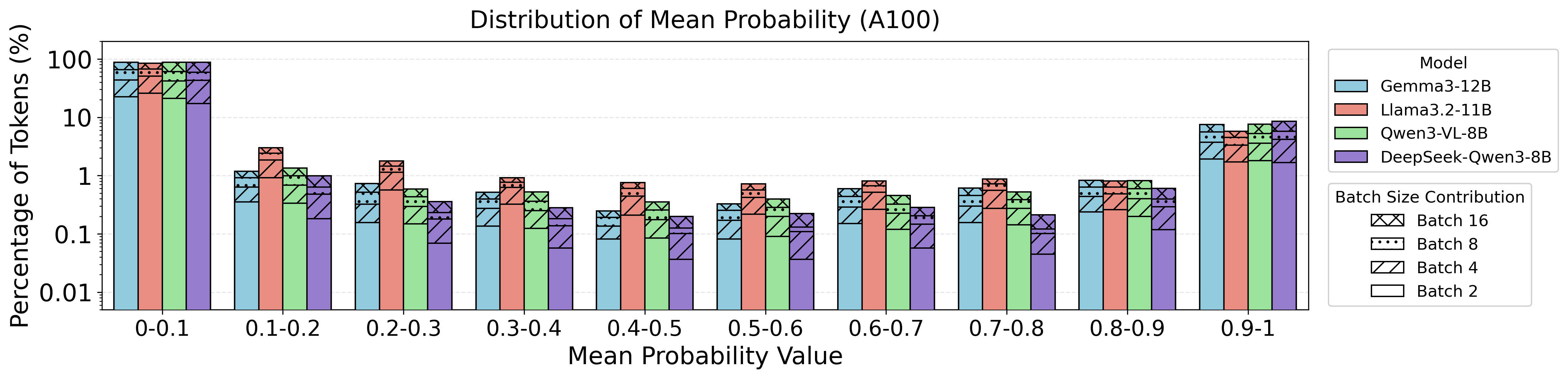}
    \includegraphics[width=0.98\textwidth]{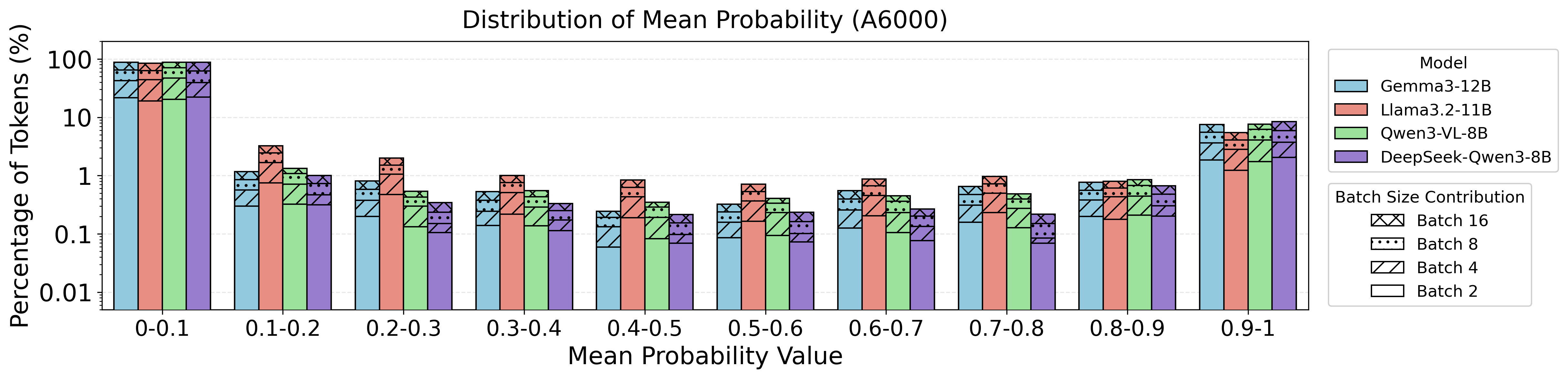}
    \includegraphics[width=0.98\textwidth]{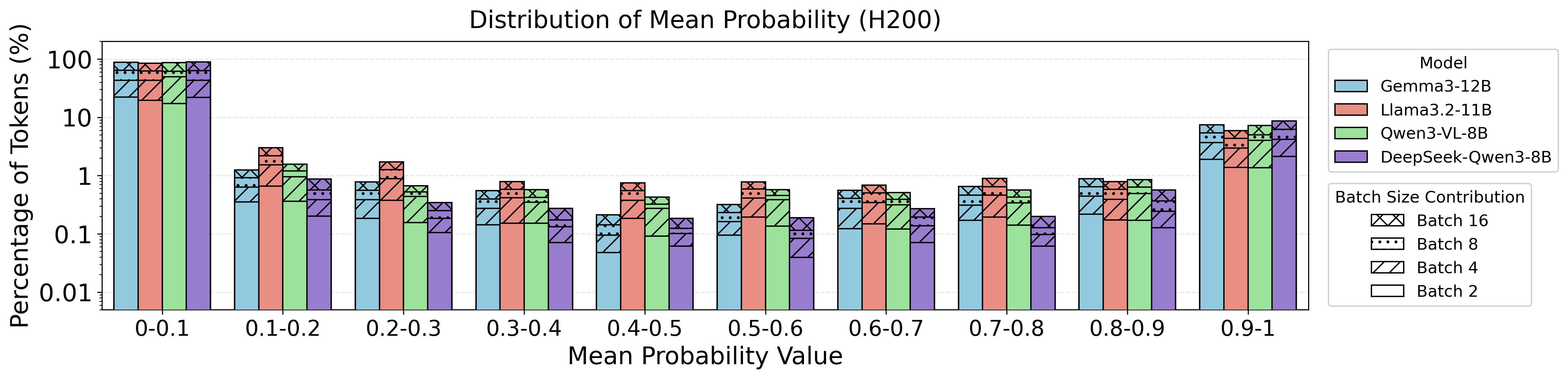}
    \includegraphics[width=0.98\textwidth]{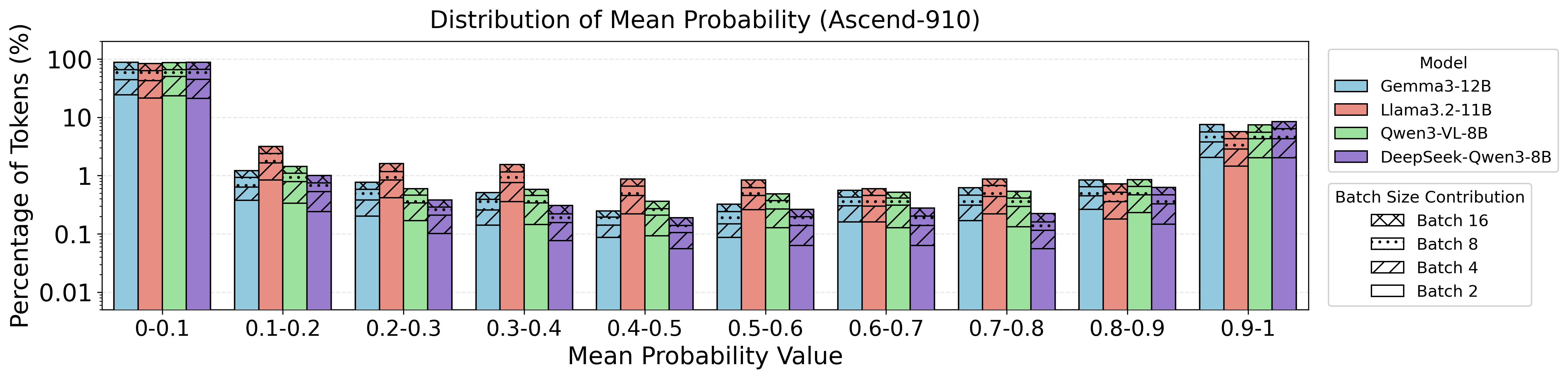}

    \caption{Distribution of average token probability for different batch sizes and models across different GPUs: NVIDIA A100 (top), NVIDIA A6000, NVIDIA H200, Huawei Ascend-910}
    \label{fig:probability-histogram}
\end{figure}

To better understand the token probabilities, in Figure \ref{fig:probability-histogram}, we take a closer look at the top 10 token probability values by examining the distribution of their average probability value. It can be seen that most of the values, roughly 90\%, are concentrated around zero (in the 0 to 0.1 range). This is because in many cases only one or at most a few tokens have sizeable probabilities, while the rest are negligible. This means that on the top-10 tokens most have low values. In fact, for many token predictions, there is only a single token with relevant probability that is close to one and the rest are close to zero, for example, when a token completes or continues a word made of several tokens. These results combined with the previous distribution of ranges for the variations of token probability in Figure \ref{fig:range-SD-histogram} suggest that using top-p sampling \cite{holtzman2019curious} can mitigate the effects of nondeterminism, because it filters out tokens whose probability was close to zero (most of the tokens) and that has increased due to the nondeterminism variations. However, as we will see next, that is not the case when we look at the effect of nondeterminism versus token probability.  

\subsection{Analysis as a function of token probability}

\begin{figure}[ht]
    \centering
    \includegraphics[width=0.98\textwidth]{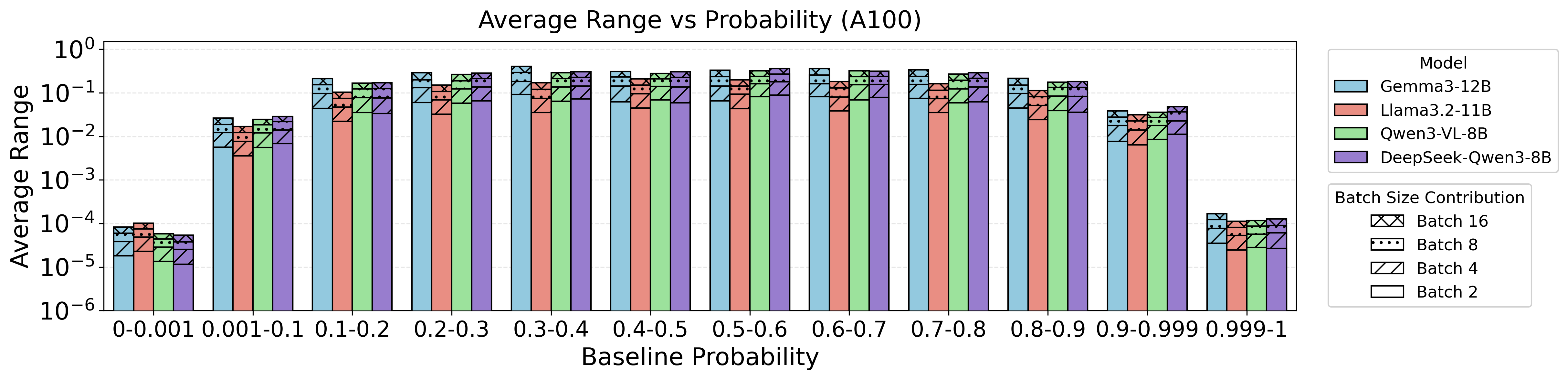}
    \includegraphics[width=0.98\textwidth]{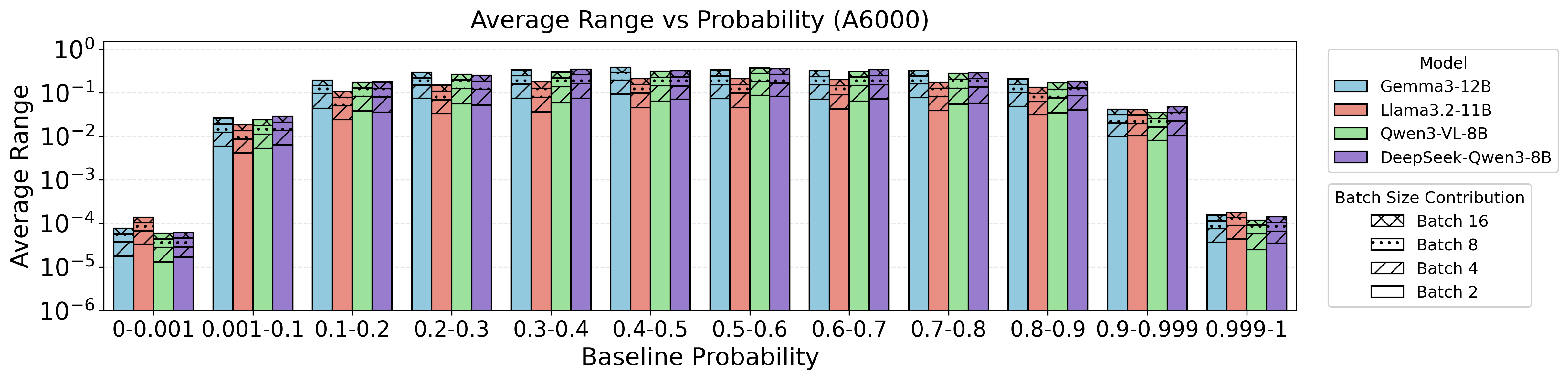}
    \includegraphics[width=0.98\textwidth]{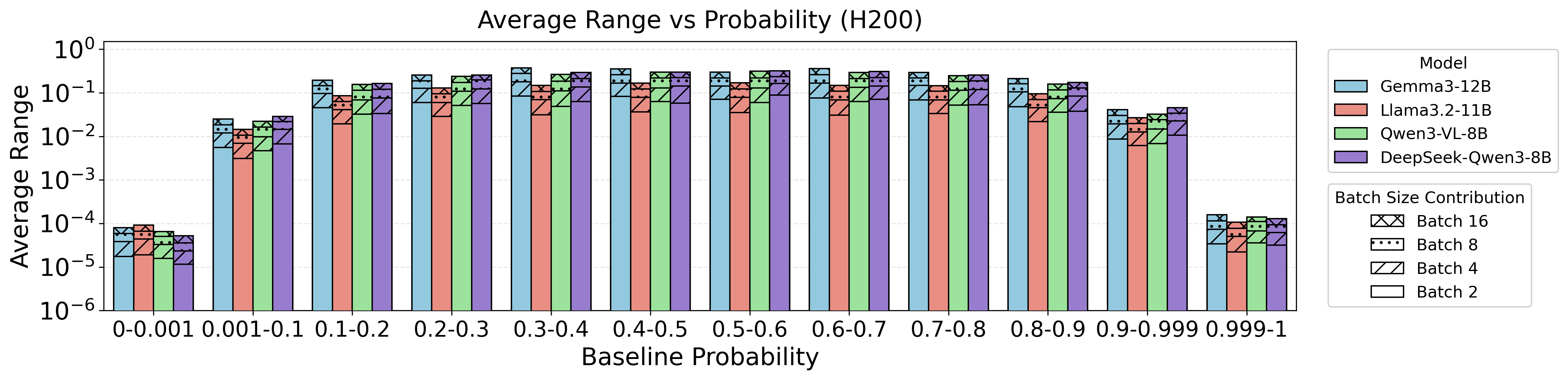}
    \includegraphics[width=0.98\textwidth]{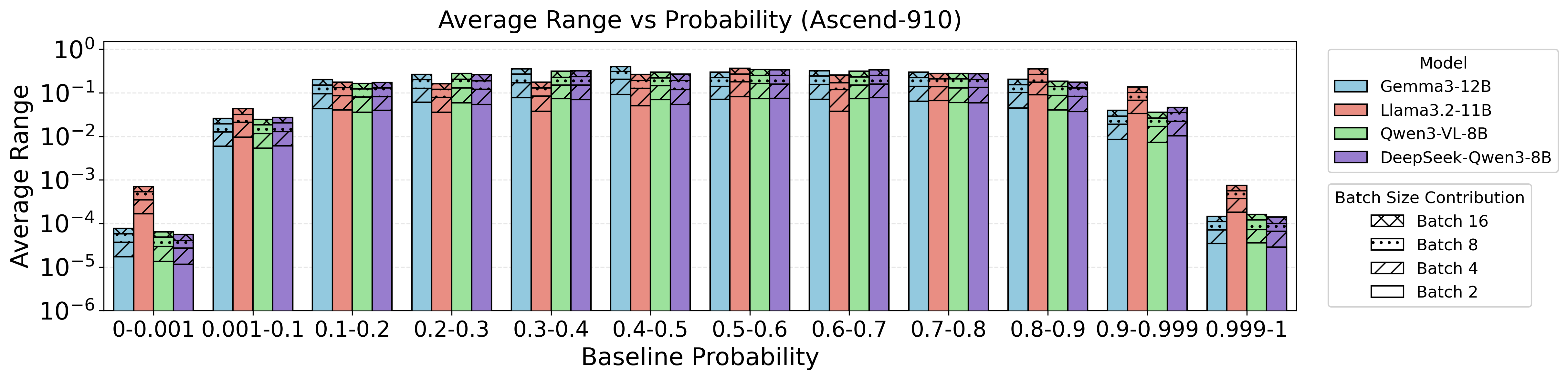}

    \caption{Average Range $R_j$ of variations in token probability due to nondeterminism for different batch sizes and models as a function of the token probability for different GPUs: NVIDIA A100 (top), NVIDIA A6000, NVIDIA H200, Huawei Ascend-910}
    \label{fig:Ranges-vs-GPUs}
\end{figure}

\begin{figure}[ht]
    \centering
    \includegraphics[width=0.98\textwidth]{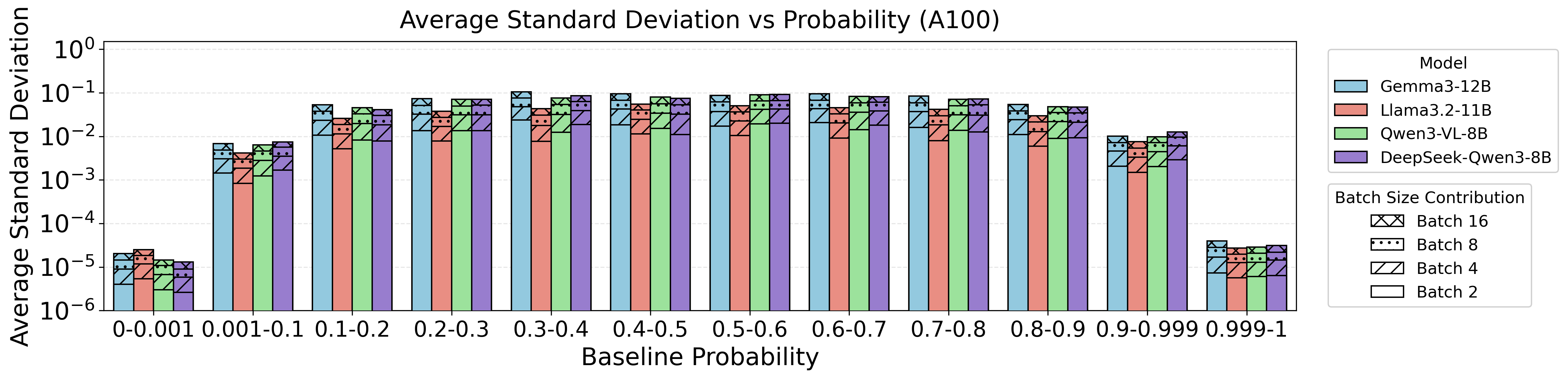}
    \includegraphics[width=0.98\textwidth]{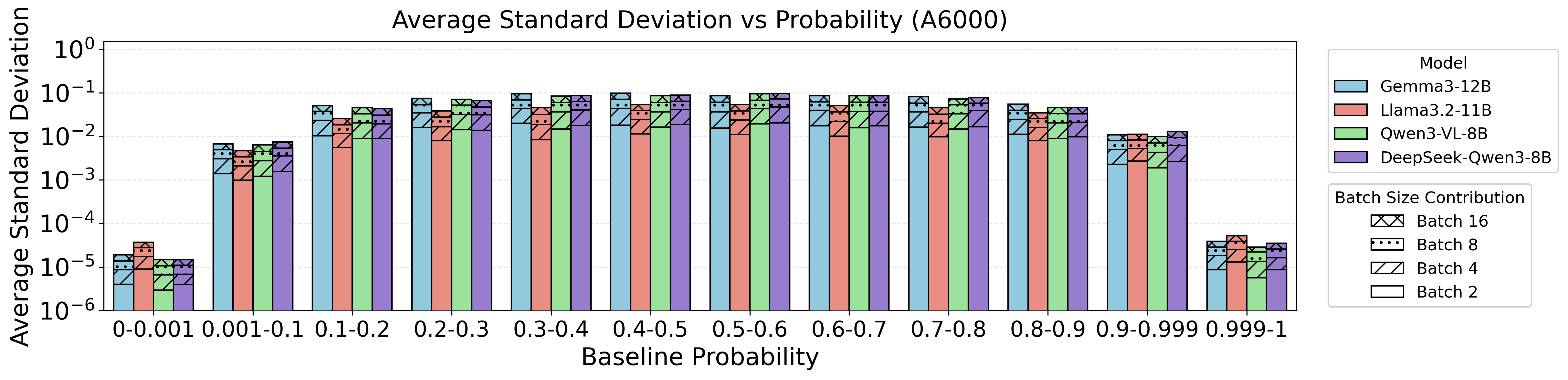}
    \includegraphics[width=0.98\textwidth]{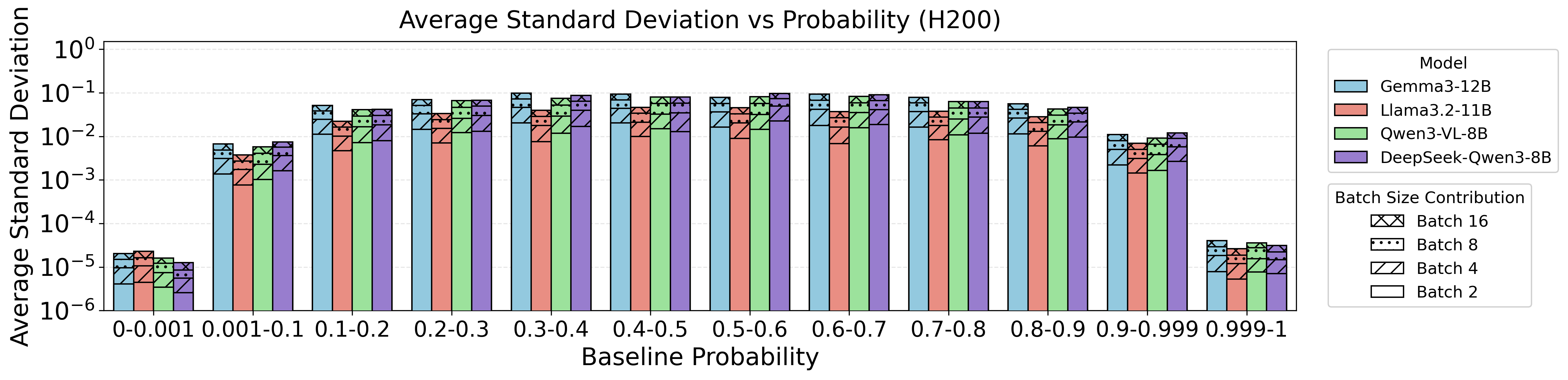}
    \includegraphics[width=0.98\textwidth]{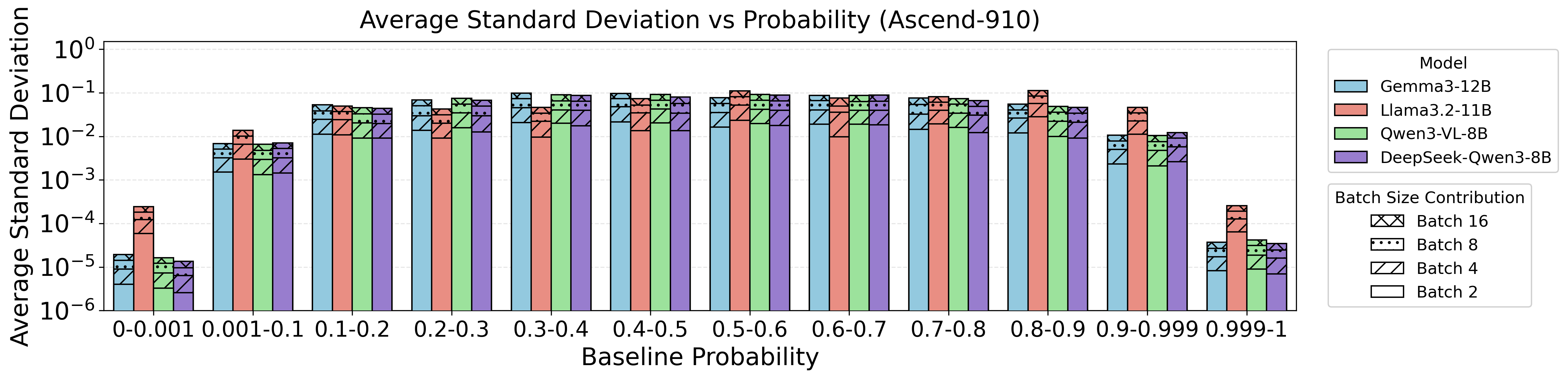}

    \caption{Standard deviation $\sigma_j$ of the variations in token probability due to nondeterminism for different batch sizes and models as a function of the token probability for different GPUs: NVIDIA A100 (top), NVIDIA A6000, NVIDIA H200, Huawei Ascend-910}
    \label{fig:Std-vs-GPUs}
\end{figure}

In Figures \ref{fig:Ranges-vs-GPUs} and \ref{fig:Std-vs-GPUs} we analyze the dependence of the impact of nondeterminism on the token probability for different batch sizes, models, and GPUs in terms of the range $R_i$ and standard deviations $\sigma_i$ of the variations observed for each token. The results show consistent trends regardless of the model, GPU or batch size: the variation is negligible when the token probability is close to 0 or 1 and takes significant values when it is in the 0.2 to 0.8 range. Therefore, a first observation is that nondeterminism is only an issue for tokens on which there are several candidate tokens with significant probabilities. 

\begin{figure}[ht]
    \centering
    \includegraphics[width=0.98\textwidth]{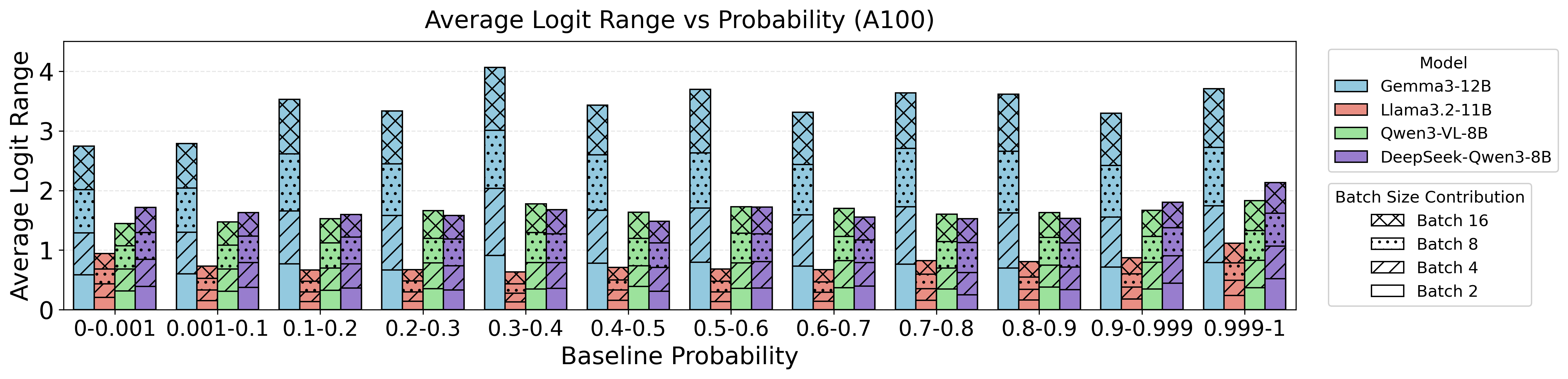}
    \caption{Average Range $R_j$ of variations of the logits due to nondeterminism for different batch sizes and models as a function of the token probability for NVIDIA A100.}
    \label{fig:Ranges-logits}
\end{figure}

To understand this behavior, we looked at the variations of the logits $z_i$ before applying the softmax normalization to obtain the probabilities \cite{goodfellow2016deep}. The results show similar variations across all the token probability ranges. An example is shown in Figure \ref{fig:Ranges-logits} for the A100 GPU. It can be seen that the values are similar for all probability values, including those close to 0 or 1. This suggests that softmax normalization:

\[
p_i = \mathrm{softmax}(z_i) = \frac{e^{z_i}}{\sum_{j=1}^{n} e^{z_j}}.
\]

is reducing the variations for tokens with probability close to 0 or 1. Since softmax is a nonlinear and saturating function, its sensitivity to variations in \(z_i\) depends on the relative differences between logits. This explains the small probability variations when \(p_i \approx 0\) or \(p_i \approx 1\) and the large variations when \(p_i\) is in the mid range (e.g., \(0.2 \le p_i \le 0.8\)). 

Let us consider the different values of the probabilities:

\paragraph{Case 1: Probability close to 0}
If a token has a probability close to 0, this implies that other tokens \(k\) have significantly larger logits, i.e., \(z_k \gg z_i\). Therefore:

\[
e^{z_i} \ll \sum_{j=1}^{n} e^{z_j} \approx \sum_{k \neq i} e^{z_k},
\]

and the softmax probability

\[
p_i = \frac{e^{z_i}}{\sum_{j=1}^{n} e^{z_j}}
\]

is close to zero and almost insensitive to small variations in \(z_i\).

\paragraph{Case 2: Probability close to 1}
If a token has a probability close to 1, this means that

\[
z_i \gg z_j \quad \forall j \ne i.
\]

Then:

\[
p_i = \frac{e^{z_i}}{e^{z_i} + \sum_{j \ne i} e^{z_j}} \approx \frac{e^{z_i}}{e^{z_i}} = 1.
\]

Small perturbations \(\delta z_i\) result in:

\[
p_i' = \frac{e^{z_i + \delta z_i}}{e^{z_i + \delta z_i} + \sum_{j \ne i} e^{z_j}} \approx 1,
\]

so the probability stays close to one and changes very little.

\paragraph{Case 3: Probability in the mid range}
When the token probability takes a moderate value (e.g., \(0.2 \le p_i \le 0.8\)), several logits are of comparable magnitude:

\[
z_i \sim z_j \quad \text{for some } j \neq i.
\]

\begin{figure}[ht]
    \centering
    \includegraphics[width=0.6\textwidth]{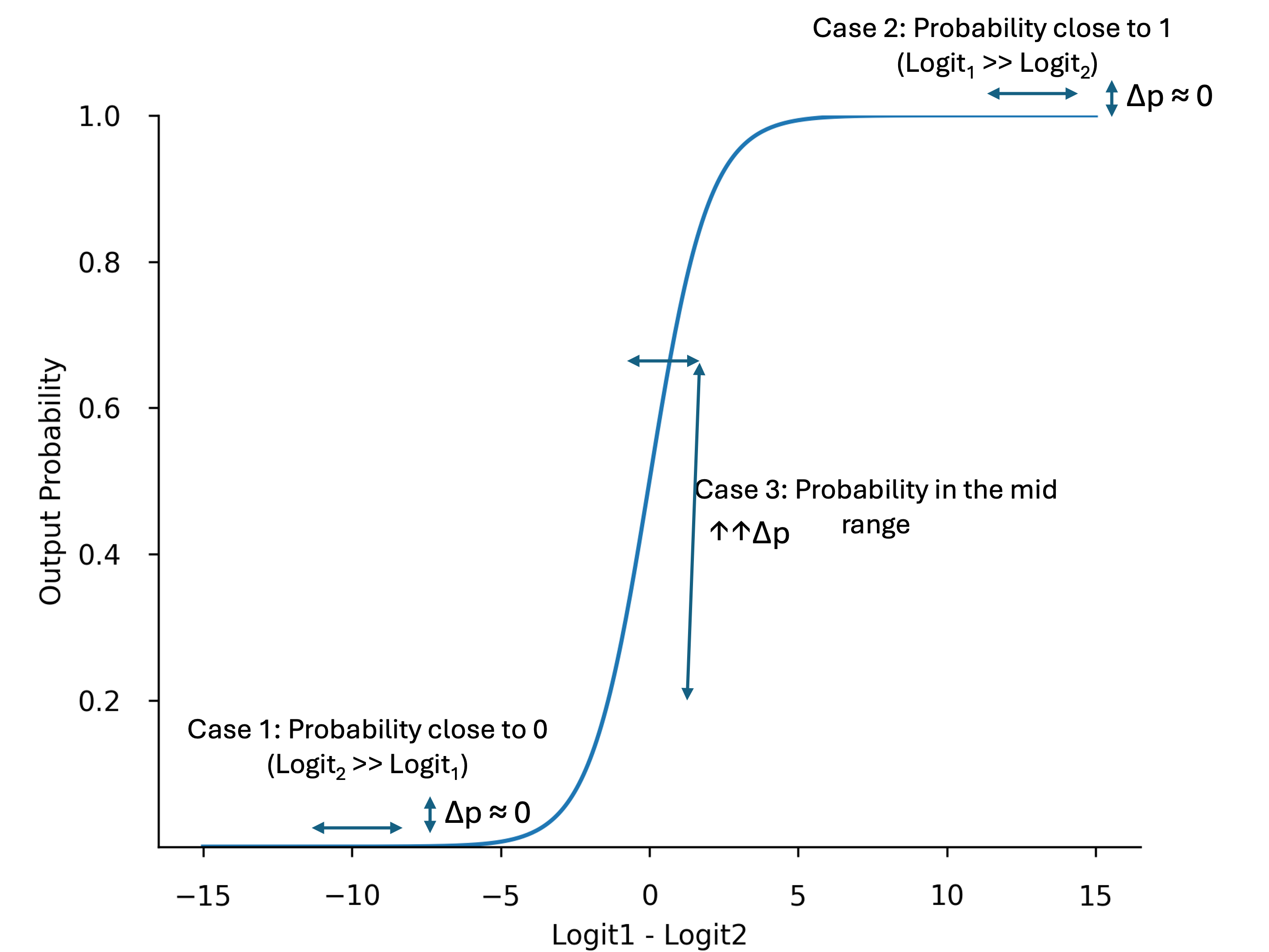}
    \caption{Effect of Softmax in non-determinism. The probability of the first token in a two-token vocabulary calculated by Softmax is identical to the probability of the positive class calculated by the Sigmoid function on the difference of the logits.}
    \label{fig:sigmoid}
\end{figure}

Here, small changes \(\delta z_i\) can significantly shift the softmax probability because both the numerator and the denominator are of similar scale, leading to greater variations in the resulting probability. Therefore, the softmax function acts as a nonlinear amplifier in the central range and a suppressor near the boundaries. Figure \ref{fig:sigmoid} represents this effect visually in the case of a two-token dictionary where the token probability (Softmax function) matches the Sigmoid function of the difference of the logit values. 

$$p_1 = \text{Softmax}([z_1, z_2])_1 = \frac{e^{z_1}}{e^{z_1} + e^{z_2}} = \sigma(z_1 - z_2)$$

This explains why nondeterminism is only an issue for tokens with several competing candidates of similar scores. This behavior is consistent across models, batch sizes, and GPUs because it is inherent to the mathematical properties of softmax rather than hardware specific effects.

Let us now consider the impact of batch size. It can be seen that, both in range and in standard deviation, the variations increase with batch size for all models and GPUs. This trend suggests that larger batch sizes introduce greater variability in computations, potentially due to memory contention and differences in parallel execution patterns. Consequently, while increasing batch size can improve throughput and GPU utilization, it also appears to amplify fluctuations in the results, which should be taken into account when selecting batch sizes.


Looking at the models, the results for the four models evaluated are very similar, with Llama3.2-11B having slightly lower variations. This suggests that the impact of nondeterminism is comparable across open-weight models of similar scale, even when they come from different vendors. In other words, despite differences in architecture refinements or training data, models of roughly the same size appear to exhibit similar sensitivity to nondeterministic effects, with only minor deviations as those seen in Llama3.2-11B.

\begin{figure}[ht]
    \centering
    \includegraphics[width=0.98\textwidth]{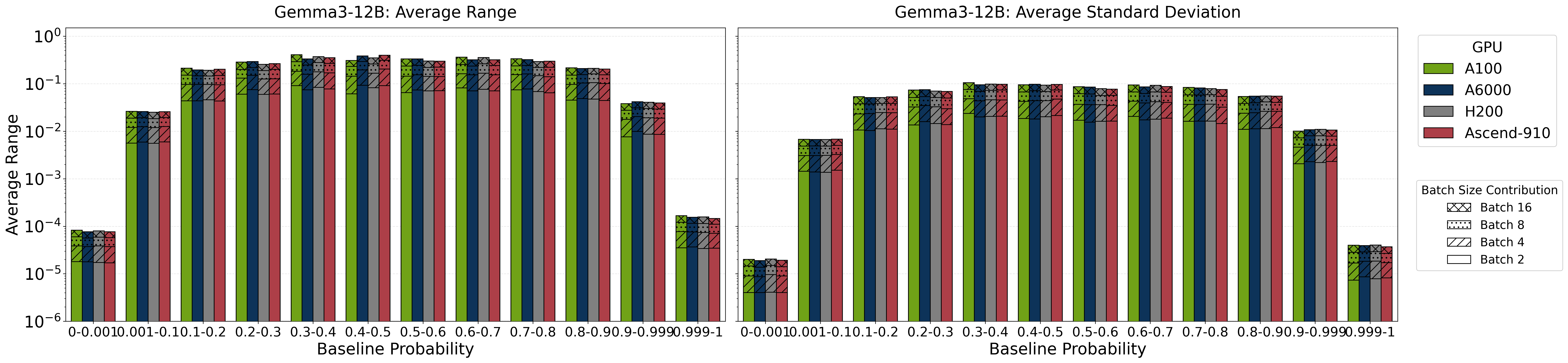}
    \includegraphics[width=0.98\textwidth]{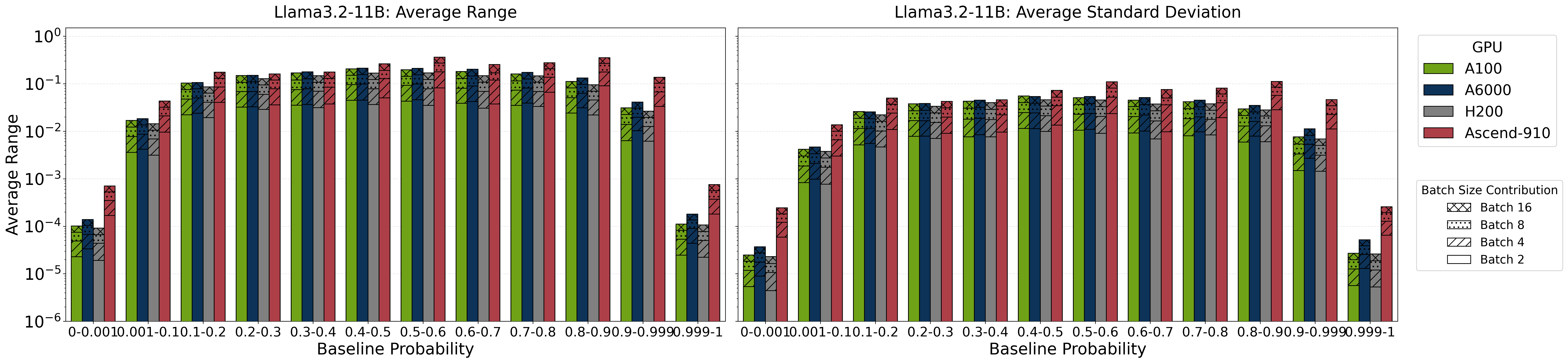}
    \includegraphics[width=0.98\textwidth]{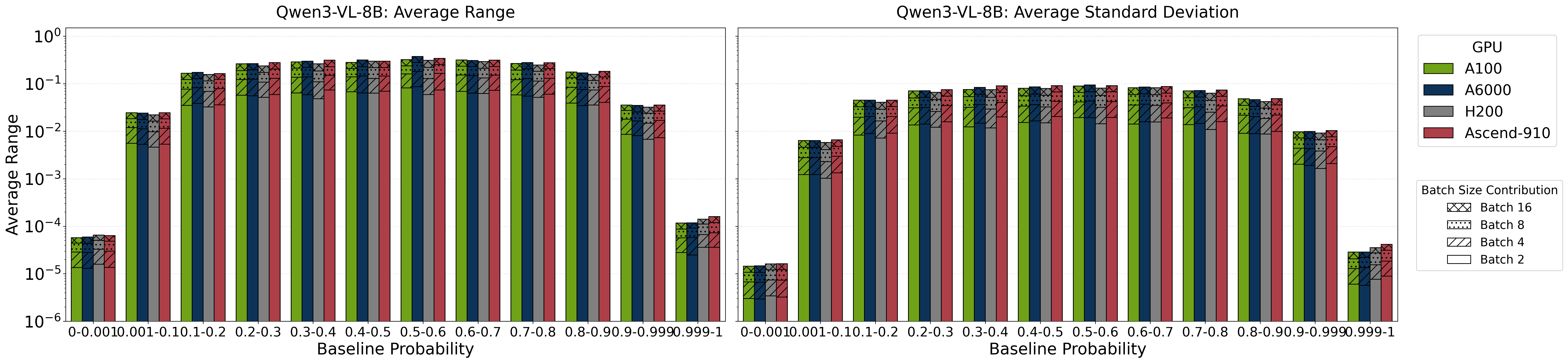}
    \includegraphics[width=0.98\textwidth]{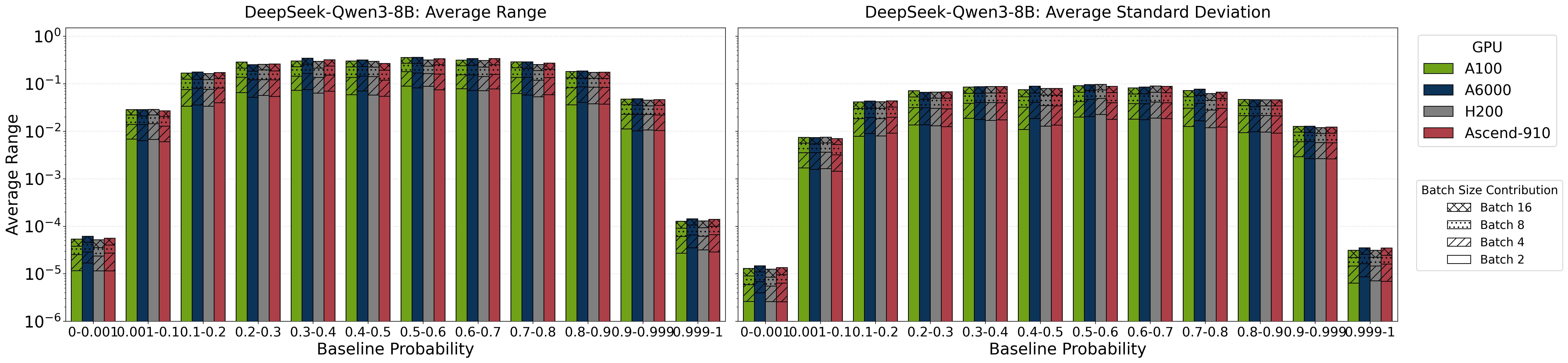}

    \caption{Standard deviation $\sigma_j$ and range $R_j$of the variations in token probability due to nondeterminism for different batch sizes and GPUs as a function of the token probability for different LLMs}
    \label{fig:range-std-perModel}
\end{figure}

Finally, to compare the different hardware platforms, in Figure \ref{fig:range-std-perModel} we plot the results per model. It can be seen that both ranges and standard deviations are almost the same for the four GPUs. Therefore, it seems that the nondeterminism effects are roughly the same for a given models and floating-point format in current GPUs. This is also an interesting result and suggests that the reordering of the operations due to parallel execution of prompts is similar in different implementations. 

\subsection{Impact of model size}

\begin{figure}[ht]
    \centering
    \includegraphics[width=0.98\textwidth]{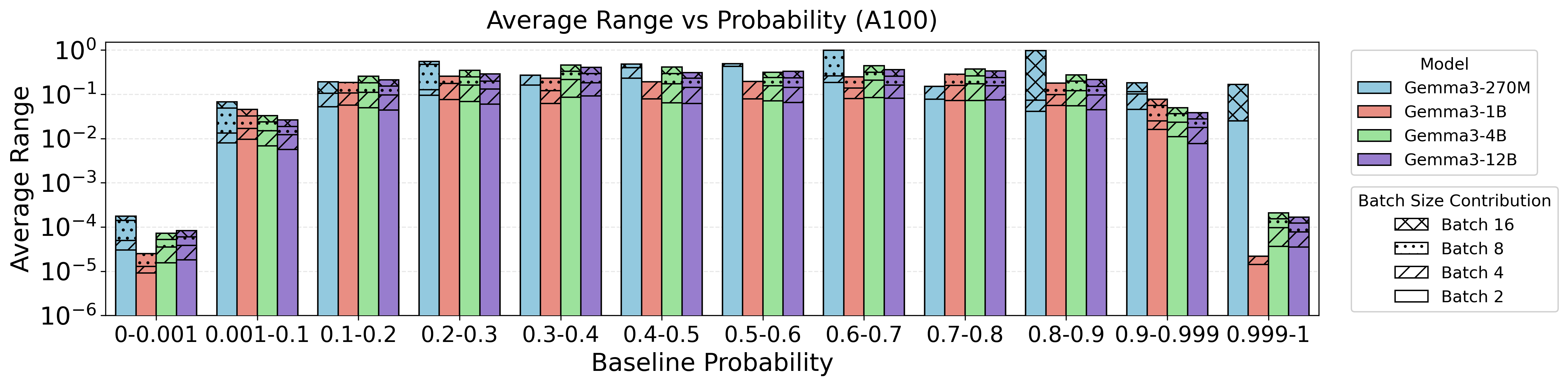}
    \includegraphics[width=0.98\textwidth]{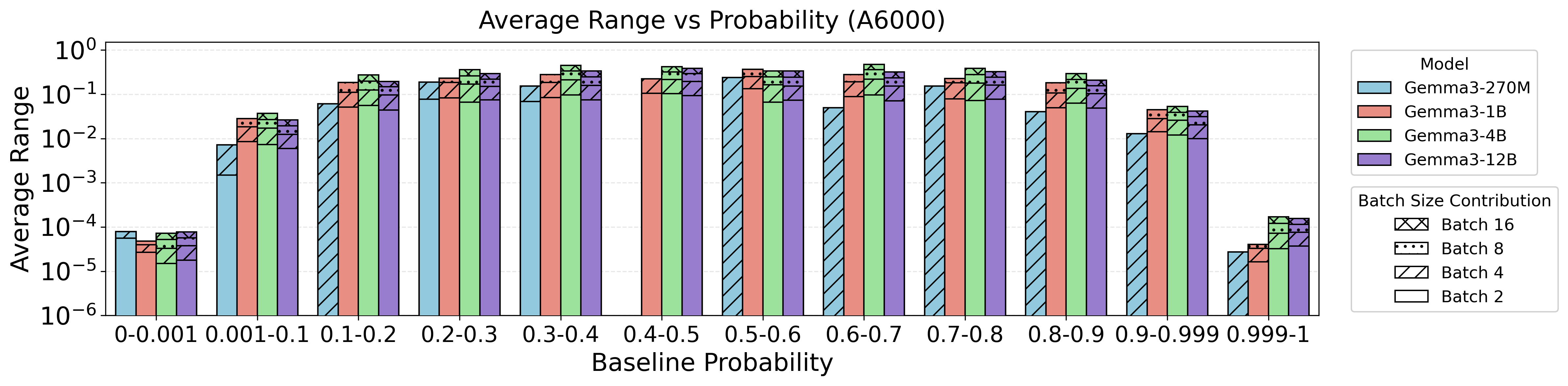}
    \includegraphics[width=0.98\textwidth]{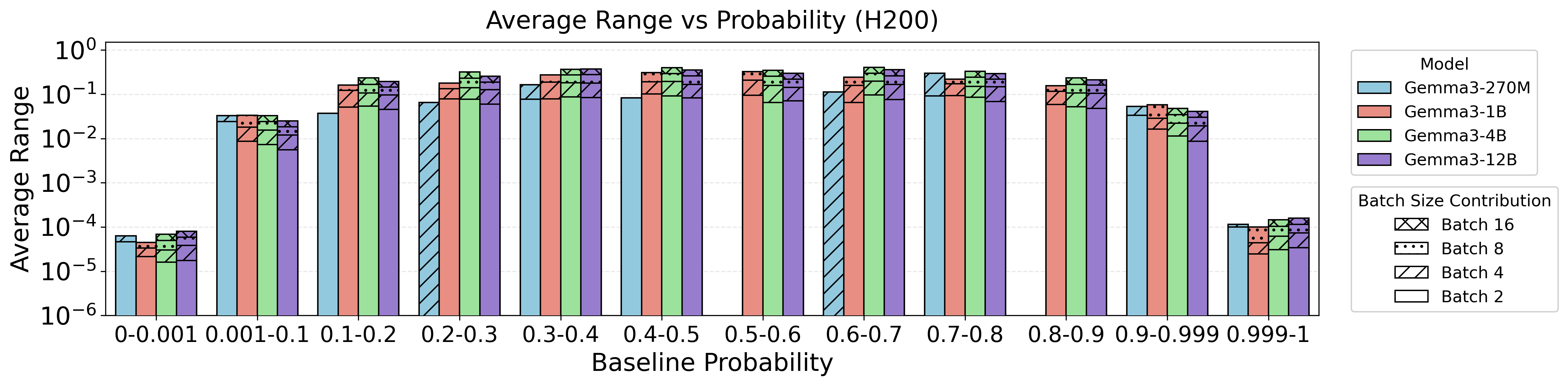}
    \includegraphics[width=0.98\textwidth]{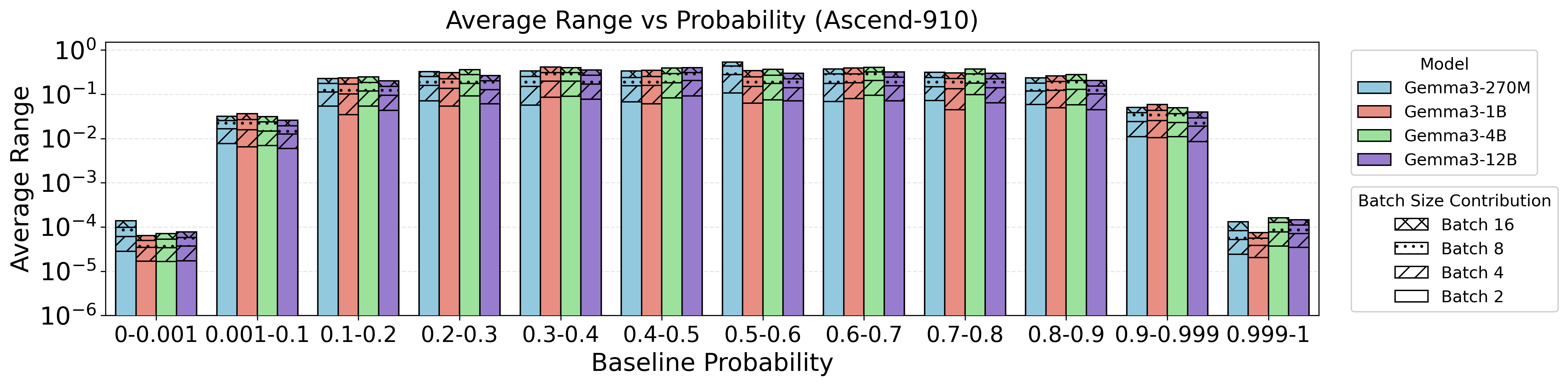}
    \caption{Average Range $R_j$ of variations due to nondeterminism for different batch sizes and models as a function of the token probability for different GPUs: NVIDIA A100 (top), NVIDIA A6000, NVIDIA H200, Huawei Ascend-910}
    \label{fig:Ranges-vs-GPUs-sizes}
\end{figure}

\begin{figure}[ht]
    \centering
    \includegraphics[width=0.98\textwidth]{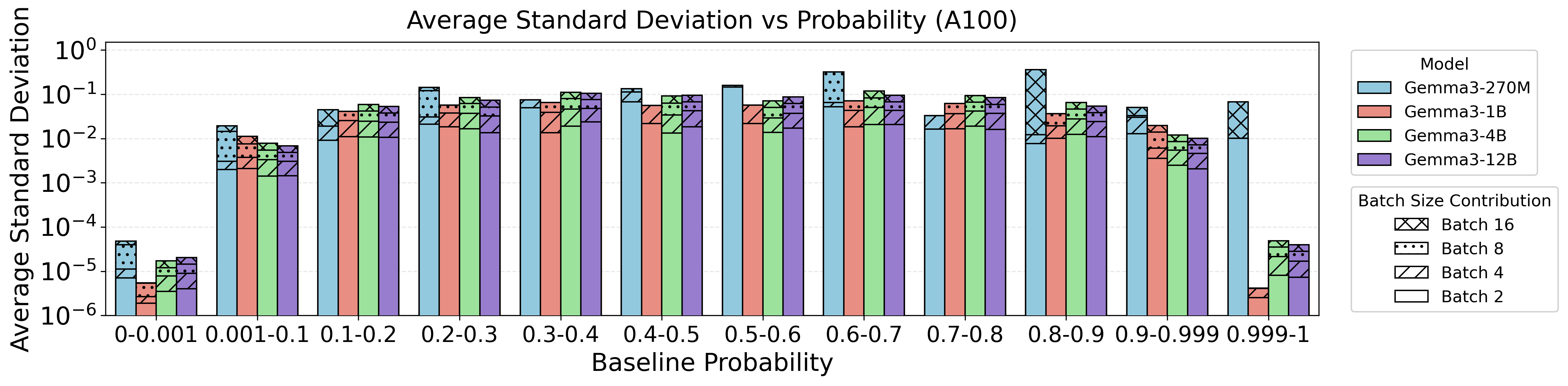}
    \includegraphics[width=0.98\textwidth]{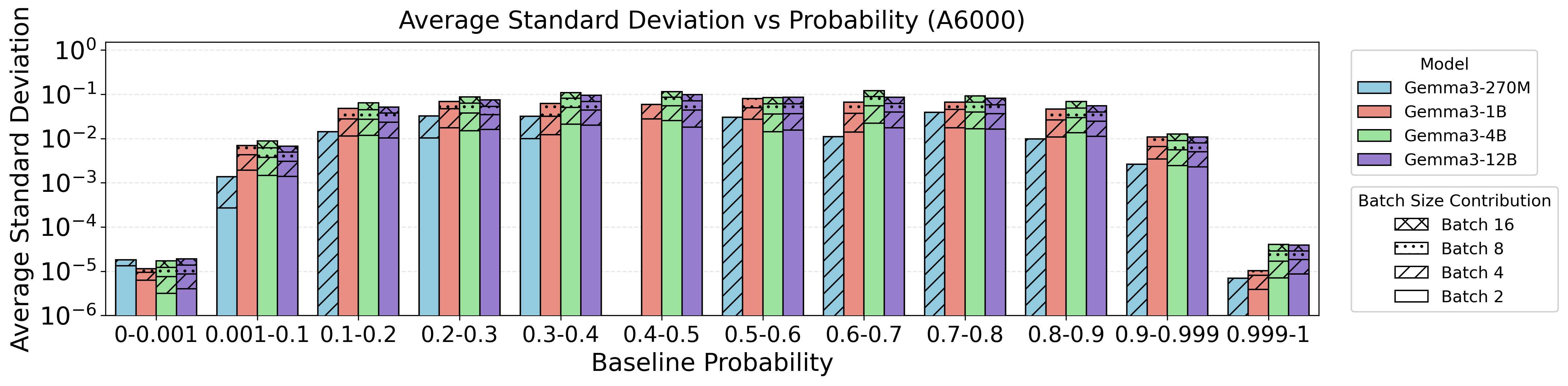}
    \includegraphics[width=0.98\textwidth]{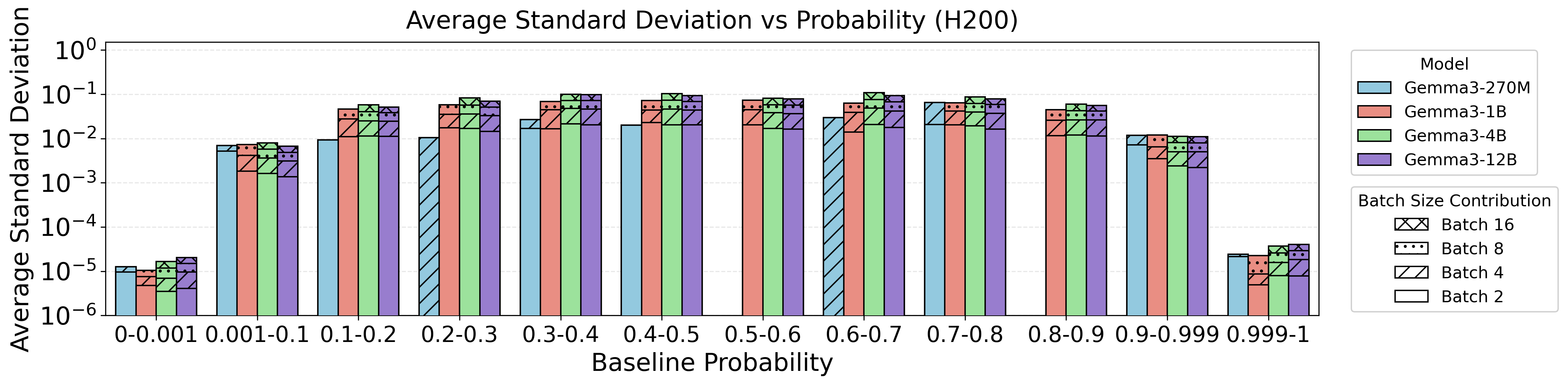}
    \includegraphics[width=0.98\textwidth]{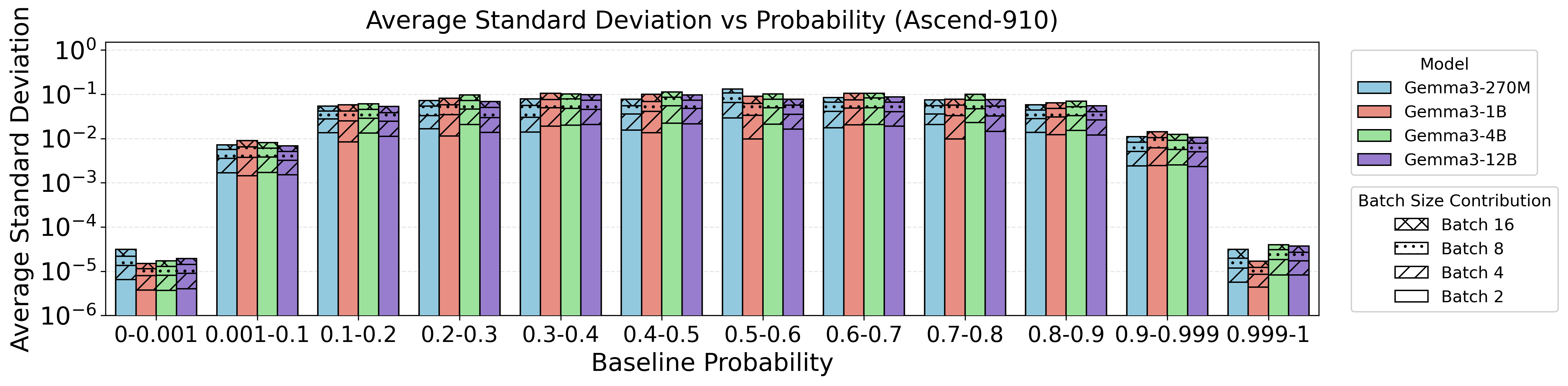}

    \caption{Standard deviation $\sigma_j$ of the variations due to nondeterminism for different batch sizes and models as a function of the token probability for different GPUs: NVIDIA A100 (top), NVIDIA A6000, NVIDIA H200, Huawei Ascend-910}
    \label{fig:Std-vs-GPUs-sizes}
\end{figure}

Another relevant parameter is the size of the model. To see if it has any effect on the level of nondeterminism, we evaluated four versions\footnote{The models evaluated were: \url{https://huggingface.co/google/gemma-3-270m-it},\url{https://huggingface.co/google/gemma-3-1b-it}, \url{https://huggingface.co/google/gemma-3-4b-it},\url{https://huggingface.co/google/gemma-3-12b-it}} of the Gemma3 model with sizes of 270M, 1B, 4B and 12B, covering a range of close to two orders of magnitude in model size. The models are from the same family to minimize differences due to model architecture or design in an effort to concentrate on the impact of model size. The results are presented in Figures \ref{fig:Ranges-vs-GPUs-sizes} and \ref{fig:Std-vs-GPUs-sizes}. It can be observed that the nondeterminism effects are similar for all sizes, with the same range and standard deviation trends; specifically, they are smaller for probabilities close to 0 or 1 and similar values for the rest of the probabilities. Interestingly, the absolute magnitude of these variations is also similar for the four model sizes, indicating that model size does not appear to significantly influence nondeterministic behavior in token probability outputs. 

Finally, it is worth noting that the smallest model, with only 270 million parameters, occasionally produced nonsensical responses in some runs. These outputs introduce additional noise into the analysis, as they contain no tokens in common with the normal responses, making it impossible to meaningfully compare token probabilities in those cases.

\section{Discussion and Limitations}
\label{sec:Discussion}

In this work, we have, for the first time, analyzed in detail the variation introduced by nondeterminism to token probabilities. This effect impacts model performance in ways that go beyond reproducibility, which has traditionally been the primary concern regarding nondeterminism. The probability variations can subtly alter the model’s behavior during text generation, influencing which tokens are selected and ultimately affecting the quality and consistency of the generated text.

Although other factors and parameters  such as the softmax temperature $T$ introduce variability in token probabilities, their effects are fundamentally different from nondeterminism. Changing $T$ systematically scales the probabilities but preserves their relative order, making higher probability tokens remain more likely than lower probability ones. In contrast, nondeterminism introduces random fluctuations that can alter the order of token probabilities, especially for intermediate probability tokens, potentially changing which token is sampled in a given run. As a result, temperature mainly affects the sharpness or flatness of the distribution, while nondeterminism introduces stochastic noise that can unpredictably modify the model’s output.

This has several implications. The first is that nondeterminism can have a substantial impact on generated text when the temperature is greater than zero, as it introduces notable variations in token probabilities, except in cases where probabilities are very close to 0 or 1. Secondly, the results suggest that different models exhibit similar levels of nondeterministic variation at the token probability level. Consequently, variations in the performance of generated text, for example, when evaluating accuracy on a benchmark, are likely driven not by differences in nondeterminism per se, but rather by differences in the underlying token probabilities or in the lengths of the responses. In terms of token probabilities, models that produce a higher proportion of probabilities near 0 or 1 are inherently less sensitive to nondeterminism. Regarding response length, models that generate shorter answers are also less likely to be affected, simply because fewer tokens are subject to stochastic variation. 

A third implication is that it may be possible to estimate the potential impact of nondeterminism from a single inference by analyzing token level probabilities, rather than needing to repeat the same inference multiple times. This approach could provide a practical and efficient way to quantify the sensitivity of models to nondeterminism and to predict its potential impact on downstream metrics or applications.

Beyond the implications of nondeterminism, this study has several limitations that can be addressed in future works:

\begin{enumerate}[a)]
    \item \textbf{Models}: only four open weight models of roughly the same size (8B to 12B) from different companies and four models of different sizes (270M to 12B) from a single company have been evaluated. Although these selections allow us to study nondeterminism effects across both similar sized models and models within the same family, the coverage remains limited. Additional vendors and a broader range of model sizes should be evaluated to confirm whether the trends observed here are consistent and generally applicable.
    \item \textbf{Single GPU}: all experiments were run on a single GPU. Nondeterministic effects may differ when models are executed across multiple GPUs, where additional sources of variation—such as inter-GPU communication, partitioning, or synchronization—can influence the variations in token probability.
    \item \textbf{GPU vendors and models}: four different GPU models from two vendors were used in the evaluation. While this provides some diversity in hardware configurations, additional GPU vendors and architectures should be evaluated to determine whether the nondeterminism effects observed are consistent across a wider range of hardware platforms. 
    \item \textbf{Prompts}: Although prompts covering different topics were used by randomly selecting questions from the MMLU benchmark for the evaluation and concurrent prompts, additional testing with a broader variety of prompt types—including, for example, longer instructions, conversational inputs, creative tasks, and domain specific prompts, would help assess whether the observed nondeterminism patterns apply in general. 
    \item \textbf{Precision format}: the results were obtained using BF16, which is commonly employed for inference. However, other numerical formats (such as FP16 or FP32) should be evaluated to determine how the choice of precision influences the nondeterminism of token probability.
    \item \textbf{Temperature}: the softmax used to compute token probabilities was applied with a temperature of 1, ensuring that no additional scaling was introduced during probability normalization. It may also be worth evaluating other temperature values ($T \ne 1$) to determine how scaling the softmax influences nondeterminism in token probabilities.
    \item \textbf{Runs}: each configuration is executed 50 times. This number seems sufficient to reveal the main variations, but performing a larger number of runs could potentially increase the observed range of values or alter the calculated standard deviation.
    \item \textbf{Language}: all prompts used in the experiments were written in English. It would be valuable to extend the analysis to other human languages, as well as to programming languages, to assess whether nondeterminism exhibits similar behaviors across different linguistic and structural domains.
\end{enumerate}

\section{Conclusion}
\label{sec:Conclusion}

Prior research on investigating the effects of nondeterminism of LLM inference has focused primarily on the reproducibility of generated text. In this work, we took a closer look and conducted the first systematic analysis of nondeterminism in LLMs at the token probability level, rather than solely at the text output level. By evaluating multiple GPUs, batch sizes, prompts, models, and run conditions, we demonstrate that probability level of nondeterminism can significantly affect generated text when the temperature is greater than zero. We further observe that different models exhibit similar levels of nondeterministic variation at the probability level with negligible impact for probabilities close to 0 or 1 and significant impact for probabilities in the range of 0.2 to 0.8. Hence, differences in benchmark performance across models are more likely due to differences in their underlying probability distributions or response lengths, rather than differences in nondeterminism itself. Models that produce probabilities concentrated near 0 or 1 are naturally less sensitive to nondeterministic effects, and models that generate shorter answers are affected less simply because fewer tokens are exposed to stochastic variations.

Building on these findings, several directions open for future research. One is to extend the analysis across a broader set of GPUs, models and prompts. A deeper theoretical understanding of the accumulation of probability perturbations across layers of the models would also help clarify their impact. Finally, exploring mitigation strategies such as hybrid precision computation based on token probabilities could provide practical solutions to improve stability in real world deployments.

\section*{Acknowledgments}

This work is supported by the FUN4DATE (PID2022-136684OB-C22) and SMARTY (PCI2024-153434) projects funded by the Spanish Agencia Estatal de Investigación (AEI) 10.13039/501100011033, by TUCAN6-CM (TEC-2024/COM-460), funded by CM (ORDEN 5696/2024) and by the Chips Act Joint Undertaking project SMARTY (Grant no. 101140087). 

\bibliographystyle{plain}

\bibliography{references}

\end{document}